\documentclass{article} 
\usepackage{iclr2026_conference,times}

\usepackage{amsmath,amsfonts,bm}

\def\eqref#1{equation~\ref{#1}}

\def\1{\bm{1}}

\DeclareMathAlphabet{\mathsfit}{\encodingdefault}{\sfdefault}{m}{sl}
\SetMathAlphabet{\mathsfit}{bold}{\encodingdefault}{\sfdefault}{bx}{n}

\usepackage{hyperref}
\usepackage{url}

\usepackage{graphicx}
\usepackage[most]{tcolorbox}
\usepackage{subcaption}
\usepackage{algorithm}
\usepackage{algorithmic}
\usepackage{multirow}

\title{Intrinsic Memory Agents: \\Heterogeneous Multi-Agent LLM Systems\\ through Structured Contextual Memory}

\author{Sizhe Yuen \& Francisco Gomez Medina \& Ting Su \\
    The Alan Turing Institute \\
    \texttt{\{syuen,fgomezmedina,tsu\}@turing.ac.uk} \\
    \And
    Yali Du \\
    King's College London \\
    \texttt{yali.du@kcl.ac.uk}
    \And
    Adam J. Sobey \\
    University of Southampton, The Alan Turing Institute \\
    \texttt{ajs502@soton.ac.uk, asobey@turing.ac.uk} \\
}

\iclrfinalcopy 
\begin{document}

\maketitle

\begin{abstract}
Multi-agent systems built on Large Language Models (LLMs) show exceptional promise for complex collaborative problem-solving, yet they face fundamental challenges stemming from context window limitations that impair memory consistency, role adherence, and procedural integrity. This paper introduces Intrinsic Memory Agents, a novel framework that addresses these limitations through agent-specific memories that evolve intrinsically with agent outputs. Specifically, our method maintains role-aligned memory that preserves specialized perspectives while focusing on task-relevant information. Our approach utilises a generic memory template applicable to new problems without the need to hand-craft specific memory prompts. We benchmark our approach on the PDDL, FEVER, and ALFWorld datasets, comparing its performance to existing state-of-the-art multi-agentic memory approaches and showing state-of-the-art or comparable performance across all three, with the highest consistency. An additional evaluation is performed on a complex data pipeline design task, and we demonstrate that our approach produces higher quality designs across 5 metrics: scalability, reliability, usability, cost-effectiveness, and documentation, plus additional qualitative evidence of the improvements. Our findings suggest that addressing memory limitations through intrinsic approaches can improve the capabilities of multi-agent LLM systems on structured planning tasks.
\end{abstract}

\section{Introduction}

Recent advances in large language models (LLMs) have enabled their application as autonomous or semi-autonomous agents capable of complex reasoning and decision-making
\citep{huang2024-planning-agents}. 
Multi-agent LLM systems, where multiple LLM instances interact to solve problems collaboratively, have shown particular promise for tasks requiring diverse expertise \citep{park2023generativeagents, qian2025scalingmulti-agent}.
These systems leverage the complementary capabilities of specialized agents to address challenges that would be difficult for single-agent approaches to resolve effectively.



Despite their theoretical advantages, multi-agent LLM systems face several implementation challenges
that limit their practical effectiveness, from coordination overhead,
to the consistency in role adherence among the agents \citep{Li2024-multi-agent-survey}. 
Most critically, 
the fixed-size context windows of LLMs restrict their ability to maintain long-term 
conversational context, an issue that is exacerbated in multi-agent frameworks with
multiple agents in a single conversation. This leads to issues such as perspective inconsistency,
forgetting key requirements, and procedural drift. Current solutions such as
Retrieval-Augmented Generation (RAG) \citep{lewis2020retrieval, gao2024rag-survey}
and agentic memory approaches \citep{packer2024-memgpt, xu2025-a-mem, mem0} 
are designed for single-agent and user interaction scenarios, which do not account for
the volume of information growing with the number of agents.


To address these challenges, we introduce Intrinsic Memory Agents, a novel multi-agent architecture that uses agent-specific memories aligned with conversational objectives. 
Unlike previous approaches, our system updates memories that are specific to each agent,
ensuring heterogeneity and memories that reflect both historical context and recent developments
while preserving agent-specific perspectives. 
The intrinsic nature of memory updates, derived directly from agent outputs rather than external 
summarization, ensures unique memories that maintain consistency with agent-specific reasoning patterns
and domain expertise.
We evaluate our approach through benchmarking and through a specific data
pipeline design case study to show its practical usage. The evaluation
demonstrates that our Intrinsic Memory Agents approach yields significant improvements in conversational coherence, role consistency, and collaborative efficiency compared to conventional multi-agent implementations. 
These improvements translate to qualitative enhancements in solution quality without increasing the number of conversation turns, suggesting broad applicability across domains where multi-agent LLM systems are deployed.

The main contributions of our work are as follows:

\begin{itemize}
    \item \textbf{Intrinsic Memory Updates}: Memory updates derived from agent outputs rather than external summarization.
    \item \textbf{Agent-Specific Memory}: Independent memories maintained for each agent to preserve perspective autonomy.
\end{itemize}


\section{Related work}\label{sec:related-work}
Recent years have seen significant progress in the development of 
multi-agent systems powered by LLMs. These systems have been applied in various domains,
such as software development, scientific experimentation, gaming, and social simulation
\citep{Li2024-multi-agent-survey}. For example, in software development, multi-agent systems enable concurrent consideration of architectural design, security, user experience, and performance optimization \citep{hong2024metagpt}.
Hallucinations due to outdated knowledge or retrieval extraction
issues remains a major challenge which limits the effectiveness of multi-agent
systems \cite{huang2025-hallucination-survey}. 
The use of a shared knowledge base or memory storage is an important
aspect to maintain up-to-date, coherent and correct information among agents.

\subsection{Memory in Agent-based systems}
In agent-based systems, memory is pivotal for maintaining context, learning from historical interactions, and making informed decisions. 
As \cite{zhang2024surveymemoryllm} noted, memory supports tasks such as ensuring conversation consistency and effective role-playing for single-agent systems.
In multi-agent systems, memory facilitates coordination, communication, and collaborative problem-solving, as \citet{guo2024llm-multiagents-survey} discussed.

Memory in LLMs can be categorized under short-term memory and long-term memory. 
Short-term memory is information that fits within the model's fixed context window. Commercial LLMs such as GPT-4o \citep{openai2024-gpt4ocard} and 
Claude \citep{anthropic-claude35} are able to process large contexts of over 100K
tokens, with some models such as Gemini 2.5 Pro \citep{comanici2025-gemini25}
able to process over 1 million tokens in its context window. However, the hard limit
of the context window size remains, and increasing the context length does not
necessarily increase reasoning or learning capabilities of the LLM 
\citep{li2024-longcontextllmsstruggle}. This is because the long context can move the relevant information further away from each other in the context window. 

Long-term memory is information that persists beyond the context window or single
instance of an LLM. This information can be stored in external databases and retrieved
using RAG techniques \citep{lewis2020retrieval, gao2024rag-survey}.
Long-term memory aims to alleviate the
issue of short-term memory's limited capacity, but 
introduces other disadvantages such as retrieval noise, 
the complexity of building a retrieval system,
latency, and storage costs \citep{asai2024selfrag, yu2024-autorag}.

The limitations of context length and existing memory mechanisms are particularly pronounced in 
multi-agent settings, where the volume of information exchanged grows with the number of agents involved \cite{li2024agentsneed}. 
As multi-agent conversations extend, the probability of critical information being at a long distance or even excluded from the accessible context increases dramatically. 
This information loss undermines the primary advantage of multi-agent systems: 
The integration of diverse, specialized perspectives toward cohesive solutions \cite{he2025llmbasedmultiagentsystemssoftware}. 
This is exacerbated by current long-term memory approaches which provide a homogeneous memory for the agents, decreasing the benefits of having agents focused on a single part of the task. Our proposed approach therefore focuses on the heterogeneity of agents and their memories, ensuring that each agent maintains a memory that is uniquely relevant to their role.



\subsection{Agentic Memory}
Agentic memory offers a solution to long-term memory and limited contextual information by periodically condensing conversation history 
into concise summaries \citep{wang2025-recursivelysummarizing, chen2024-compressivememory}. 
These approaches generate sequential or hierarchical summaries that capture key decisions and insights from previous exchanges. 
Some agentic memory approaches combine with RAG approaches by storing
the summarized contexts for retrieval later in the conversation \citep{xu-2022-beyond-goldfish-memory}, 
or by storing in- and out-of-context memory in a hierarchical system
to dynamically adapt the current context \citep{packer2024-memgpt, xu2025-a-mem}.
While agentic memory methods provide better contextual integration than pure retrieval approaches,
they frequently lose critical details during the condensation process.
Furthermore, the undirected and unstructured nature of general summarization often fails to preserve role-specific perspectives and specialized knowledge that are essential to effective multi-agent collaboration.


Our proposed Intrinsic Memory Agents similarly uses an agentic memory approach
to summarize and store information. Unlike existing approaches, we introduce
heterogeneous memory for each agent in the multi-agent system to maintain
specialized roles in collaborative tasks, and apply a templated approach to each
agent ensuring cohesive memory throughout. This addresses the limitations of existing
memory mechanisms by ensuring that each agent maintains its own memory, 
reflecting both historical context and new information while maintaining 
heterogeneous agent-specific perspectives and expertise.

\section{Intrinsic Memory Agents}\label{sec:intrinsic-memory}

The various agentic memory approaches are all designed in single-agent
scenarios to remember crucial details when interacting with an end-user.
Due to the multi-turn long conversations between agents, a direct
implementation of single-agent agentic memory becomes complicated and resource-intensive,
with each agent requiring retrieval systems and distinct contextual updates.

\begin{figure*}[!tbh]
    \centering
    \includegraphics[width = 0.8\linewidth]{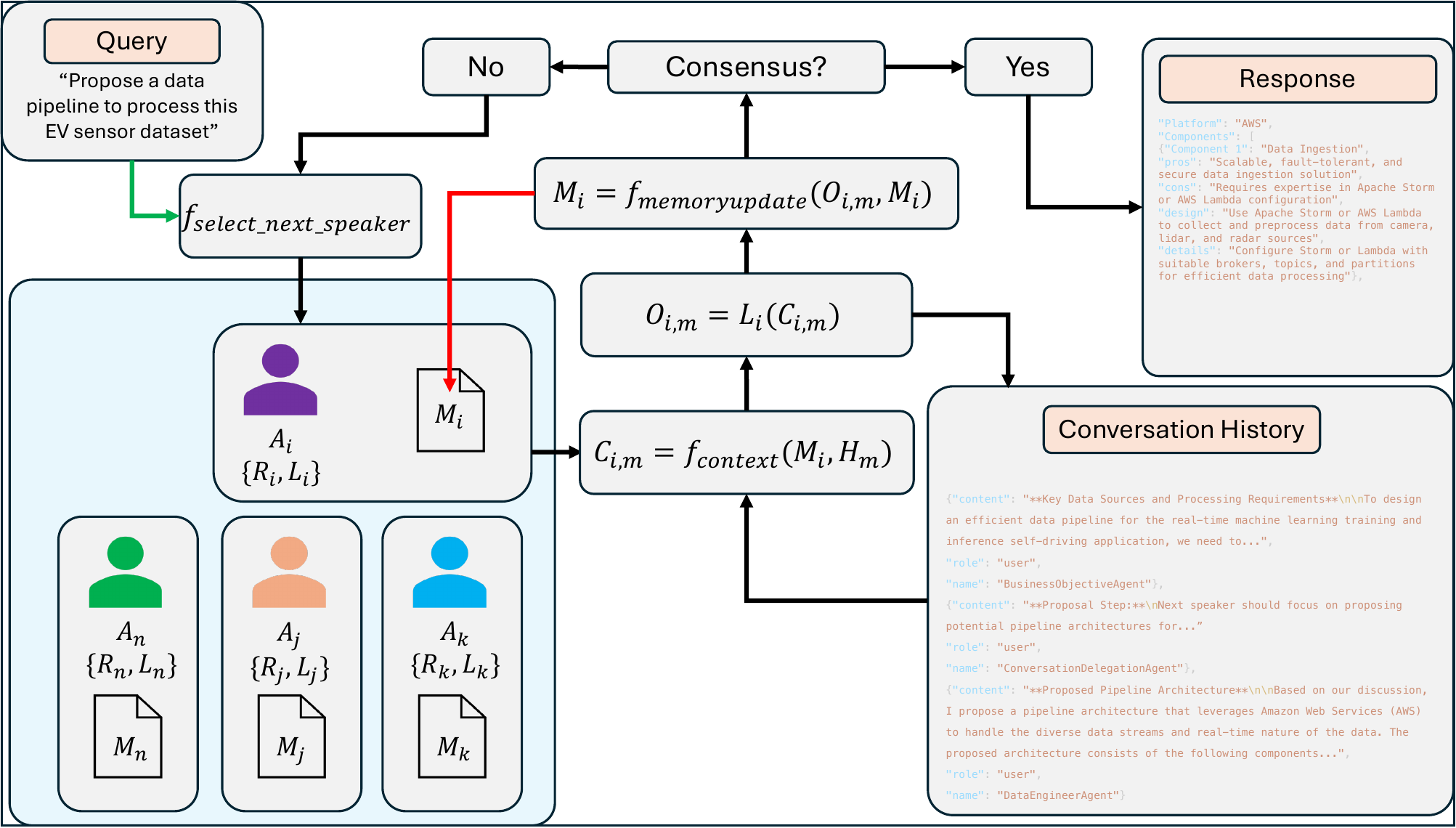}
    \caption{Intrinsic Memory Agents Framework. For $n$ agents and $m$ conversation turns, each agent $A_n$ contains its own role description $R_n$ and language model $L_n$. Its memory $M_{n,m}$ is updated based on the input context $C_{n,m}$ and output $O_{n,m}$. 
    }
    \label{fig:intrinsic_memory_agents_framework}
\end{figure*}

We propose Intrinsic Memory Agents, a framework for multi-agent LLM systems that maintains agent-specific memories aligned with conversational objectives. Figure \ref{fig:intrinsic_memory_agents_framework} illustrates the architecture of our Intrinsic Memory Agents framework. In this approach a query is made by the user, the first agent makes a comment based on its role description, the conversation is updated, followed by a memory update for the agent that commented, there is a check for consensus and the cycle starts again. The context in this case is made up of both the agent's intrinsic memory and the conversation, meaning that as the conversation continues the agents increasingly diverge in their interpretation of that context.

\subsection{Framework Definition}
Let us define the multi-agent system $\mathcal{A} = \{A_1, A_2, ..., A_N\}$ consisting of $N$ agents. Each agent $A_n = \{R_n, M_n, LLM_n\}$ is characterized by a role specification $R_n$ that defines the agent's expertise domain and objectives, a memory $M_n$ that evolves throughout the conversation, and an LLM instance $LLM_n$, which may share parameters between agents.

The conversation consists of a sequence of turns $T = {t_1, t_2, ..., t_M}$ where each turn $t_m$ involves an agent selection function $\sigma(t_m) \rightarrow A_n$ that determines which agent speaks, an input context $C_{n,m}$ constructed for the selected agent, an output $O_{n,m}$ generated by the selected agent, and a memory update operation for the selected agent.

Critically, our framework separates the input context construction and memory update processes, allowing for agent-specific memory maintenance while preserving a shared conversation space.


\subsection{Memory Update mechanism}
For each agent in the system, we maintain a memory $M_n$ that evolves over time. Let $M_{n,m}$ represent the memory of agent $n$ after $m$ conversation turns. The memory update process works as follows:

Agent $A_n$ receives input context $C_{n,m}$ consisting of relevant conversation history $H_m$ and previous memory $M_{n,m-1}$,
\begin{equation}
    C_{n,m} = f_{\text{context}}(H_m, M_{n,m-1});
    \label{equa-c_nm}
\end{equation}

and agent $A_n$ generates output $O_{n,m}$ using the underlying LLM $L_n$,
\begin{equation}
    O_{n,m} = L_n(C_{n,m}).
\end{equation}

Then with the generated output $O_{n,m}$ and the previous memory $M_{n,m-1}$, we update the slot content using a memory update function,
\begin{equation}
    M_{n,m} = f_{\text{memory\_update}}(M_{n,m-1}, O_{n,m}).
\end{equation}

The memory update function $f_{\text{memory\_update}}$ is implemented as a prompted LLM operation. Specifically, for the previous memory $M_{n,m-1}$ at turn $m-1$ and agent output $O_{n,m}$ at turn $m$, the update function constructs the prompt as shown in Figure \ref{fig:memory-update-prompt}. The LLM's response to this prompt becomes the updated memory $M_{n,m}$. The context construction function $f_{\text{context}}$ presented in equation~\ref{equa-c_nm} determines what information is provided to an agent when generating a response. 
The algorithm takes the existing conversation history and agent memory,
appending both to the context and using the remaining tokens to include
the rest of the conversation history. The full algorithm pseudo-code is
displayed in the Appendix \ref{apdx:algorithms} Algorithm \ref{Alg:Context_construction}.

This algorithm prioritizes:
\begin{enumerate}
    \item The initial task description to maintain objective alignment. 
    \item The agent's structured memory to preserve role consistency.
    \item The most recent conversation turns to maintain immediate context.
\end{enumerate}

By prioritizing memory inclusion over exhaustive conversation history, the algorithm ensures that agents maintain role consistency and task alignment even when conversation length exceeds context window limitations. We conduct an ablation study on the structure of the memory template in appendix \ref{apdx:ablation}. The ablation study shows the use of a generic or dynamic LLM-generated template
shows consistently better performance compared to a hand-crafted template, which is prone to sensitivity if a poorly created template is used.

\section{Quantitative benchmarks}\label{sec:pddl-benchmark}

To evaluate our approach, we test our memory agents against the PDDL (Planning Domain Definition Language), FEVER (Fact Extraction and VERification) \cite{thorne-etal-2018-fever}, and ALFWorld \cite{ALFWorld20} numeric benchmarks. 
PDDL involves structured planning 
tasks from AgentBoard \citep{ma2024-agentboard}, where the agents generate executable 
plans for abstract problem domains, evaluating their reasoning and coordination.
FEVER is a dataset for evidence-based claim verification, requiring agents to 
retrieve and reason over textual evidence and assess a given factual claim.
Finally, ALFWorld \cite{ALFWorld20} is a text-based interactive environment which simulates
household tasks with natural language instructions and descriptions.
It tests an agent's ability to navigate and execute complex sequential actions to complete tasks.

For numerical benchmarks, we follow the same experimental methodology as G-Memory 
\citep{zhang2025-gmemory}, another memory
framework for multi-agent systems. We re-run the G-Memory framework \footnote{https://github.com/bingreeky/GMemory} as we cannot directly compare to the published G-Memory results
which were benchmarked with GPT-4o-mini as the base language model.
We chose to use the G-Memory framework as a comparison as the framework
implements a variety of existing memory architectures, allowing us to compare
our Intrinsic Memory Agents with existing architectures and benchmarks.
G-Memory uses Autogen for multi-agent simulation, matching our use of Autogen for
our architecture. We chose to use the three benchmarks to cover a range of structured planning, comprehension, and
reasoning tasks, all of which are aspects aligned with the
data pipeline case study detailed in Section \ref{sec:data-pipeline}.
We run Gemma3:12b for the numeric benchmarks
using Ollama \footnote{https://ollama.com/library} with 5 independent runs, each with their own
set seeds for reproducibility. We use a larger
model for the numeric benchmarks as initial tests on the Llama3.1:3b model found poor results for every benchmark
and memory framework. Our computational infrastructure utilizes
a high performance computing cluster with A100 GPUs, running on GNU/Linux 4.18.0-553.el8\_10.x86\_64.

\subsection{Benchmarking results}

\begin{figure}[h!]
    \centering
    \begin{tabular}{ccc}
        \includegraphics[width=0.32\textwidth]{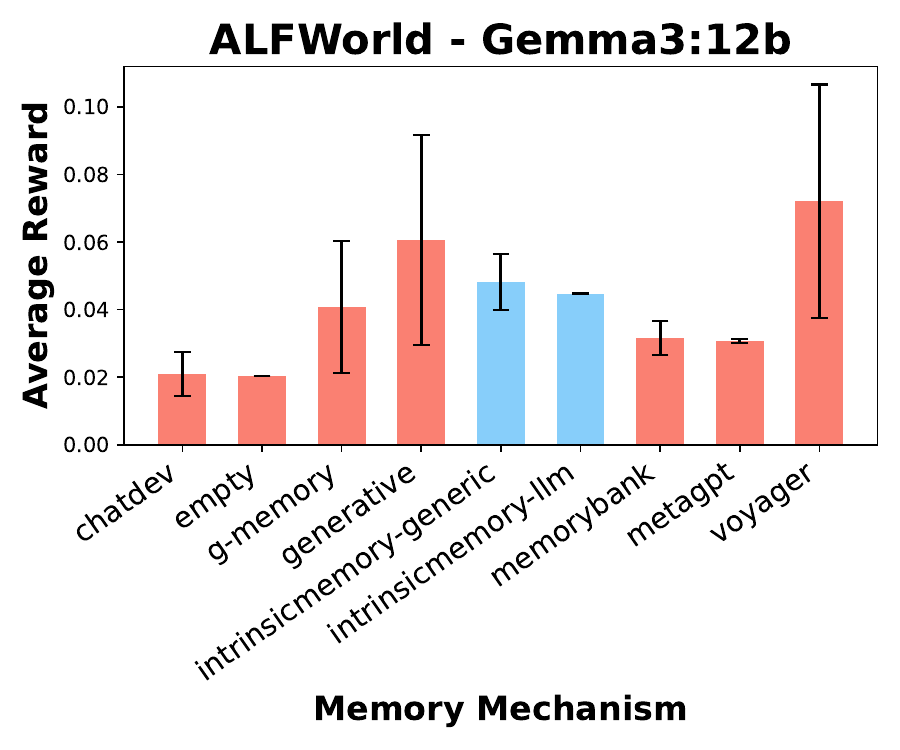} &
        \includegraphics[width=0.32\textwidth]{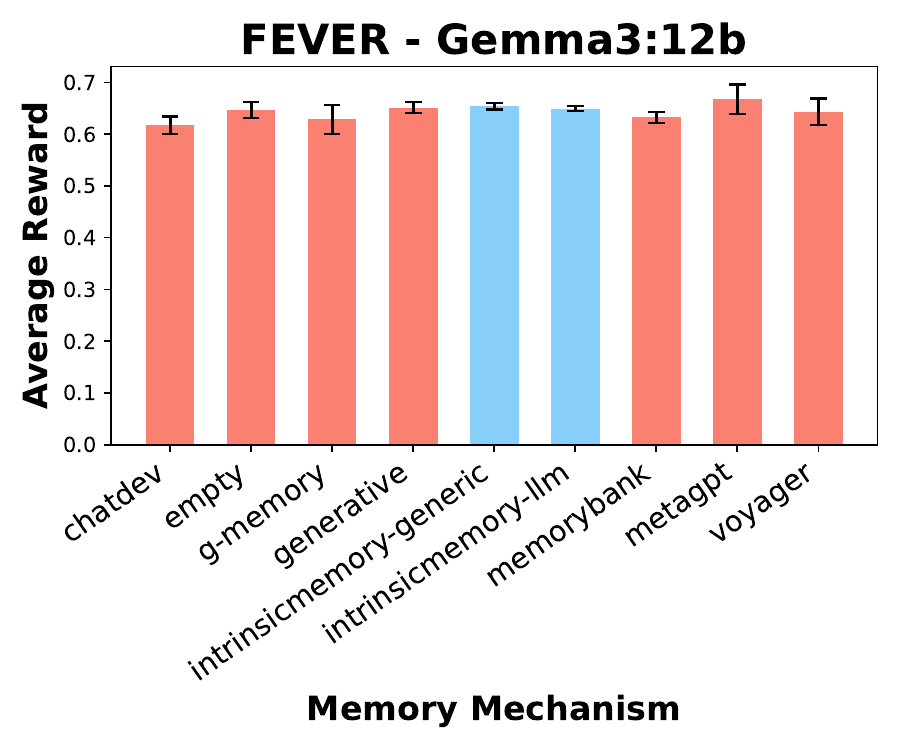} &
        \includegraphics[width=0.32\textwidth]{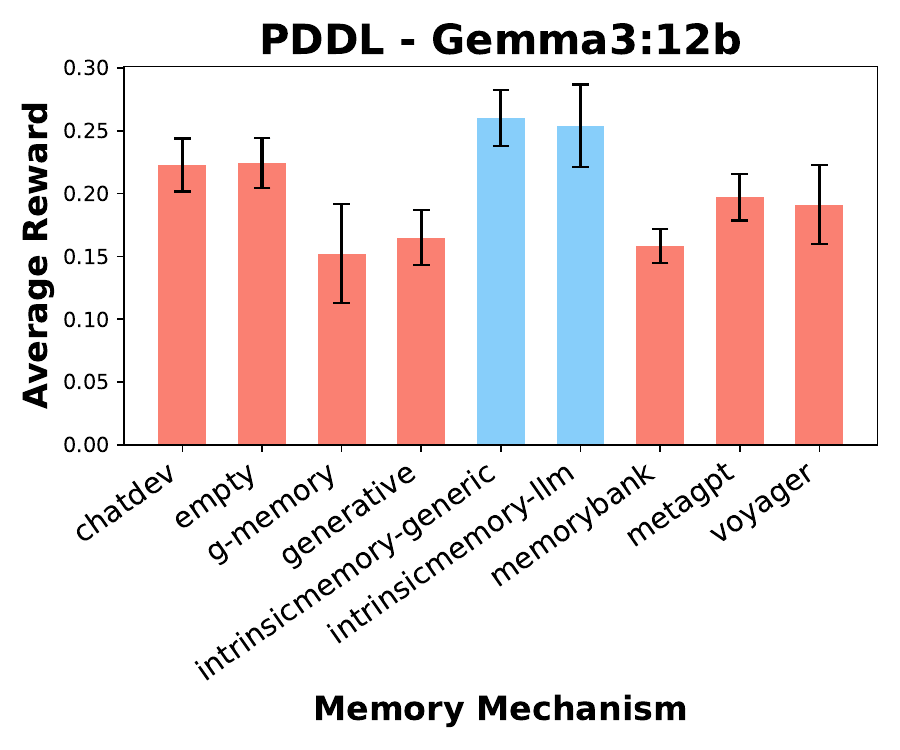} \\
    \end{tabular}
    \caption{Intrinsic Memory performance across the three benchmarks, the blue bars are our Intrinsic Memory.}
    \label{fig:benchmark-bars}
\end{figure}


Figure \ref{fig:benchmark-bars} shows the average reward of each memory system
benchmarked against ALFWorld, FEVER, and PDDL, the error bars showing the standard deviation
across 5 independent runs. While our Intrinsic Memory mechanism doesn't obtain the highest
rewards in each benchmark, there are results to indicate our approach has consistent strong 
performance compared to the other memory mechanisms.

In the ALFWorld benchmark, the Voyager and Generative approaches obtain the highest average
reward, at 0.072 and 0.061 mean reward respectively. However, they also show the highest standard deviation among all memory mechanisms, at 0.035 and 0.031 respectively, indicating their
high variability. In contrast, our method with both the generic and LLM-generated templates obtains
the 3rd and 4th best performance, at 0.048 and 0.045 mean rewards, with much lower variance at 0.0083 
and 0.0003 respectively. These results indicate our strong performance and consistency when solving the
ALFWorld set of problems. Similarly for the FEVER dataset, all
memory approaches obtain similar performance, with the best approach, MetaGPT, also showing the
highest standard deviation compared to other approaches. Our Intrinsic Memory mechanism shows
the lowest standard deviation on this dataset, with mean rewards ranked second for the generic template, and fourth for the LLM-generated template, showing more evidence of our consistency.
Finally in the PDDL benchmark, both our Intrinsic Memory approaches outperform all other
memory mechanisms, with not a significantly higher standard deviation than other approaches,
at 0.260 and 0.254 mean rewards for the generic template and LLM-generated templates respectively.
Table \ref{tbl:gemma3-results-table} in the appendix shows the mean rewards, standard deviation, and
average token counts for each benchmark and memory mechanism in detail.


The PDDL dataset are structured planning tasks, which fits the intended use
case of Intrinsic Memory for agent discussion, planning and design. As
Intrinsic Memory assigns agent-specific memory, it can more clearly distinguish
planning and actions to complete tasks. More tokens are used by Intrinsic Memory
to generate structured templates per agent per round of discussion, and is a
worthwhile trade-off in both reward score and token efficiency.
In contrast, the FEVER dataset tasks are meant for fact extraction where
reasoning plays a larger role than raw memory. We find that our Intrinsic Memory performs
just as well as other memory methods, indicating memory methods in general
are less applicable to the FEVER problems, and that our performance is in line
with other memory mechanisms. Finally in ALFWorld, the two best performing memory
mechanisms, Voyager and Generative, also have the highest standard deviation, showing
a lack of consistency compared to the Intrinsic Memory approach, where the agent-specific
memory helps maintain consistent performance compared to global or cross-trial memory
implementations.


\section{Data Pipeline Design Case Study}\label{sec:data-pipeline}

As a practical case study to evaluate our approach, we applied our memory agents to a
collaborative data pipeline design, a complex task requiring multiple perspectives. We run 10 independent outputs with eight specialized agents:

\begin{enumerate} 
    \item \textbf{Evaluation Agent (EA)} evaluates the output solutions. 
    \item \textbf{Knowledge Integration Agent (KIA)} summarizes each discussion round (e.g. after every agent has contributed at least once). 
    \item \textbf{Data Engineer Agent (DEA)} determines the data processing needs.
    \item \textbf{Infrastructure Engineer (IA)} designs the cloud infrastructure.
    \item \textbf{Business Objective Engineer (BOA)} checks against business requirements.
    \item \textbf{Machine Learning Engineer (MLE)} provides ML implementation.
    \item \textbf{Conversation Delegation Agent (CDA)} is responsible for facilitating the collaborative process.
    \item \textbf{Documentation Joining Agent (DJE)} is responsible for producing final output after consensus is reached among agents.
\end{enumerate}

The agents are tasked with designing a cloud-based data pipeline architecture through a structured process involving proposals, discussions, and consensus formation. The full prompts and task descriptions can be found in Appendix \ref{apdx:data-pipeline-prompts}.

The output requirements include a concise summary, high-level plan, resource estimates, and a structured JSON specification.

\subsection{System Configurations}\label{sec:system_configuration}

We evaluated two system configurations: First, the \textbf{Baseline System} which consists of a standard multi-agent implementation without intrinsic memory. It uses standard prompt templates for each agent role, relying exclusively on conversation history for context. 
Second is our \textbf{Intrinsic Memory System} approach with agent-specific memories. It implements agent-specific memories and updates them intrinsically based on agent outputs, and constructs context using both conversation history and agent memories.


Both systems used identical agent roles and task specifications, with Llama-3.2-3b as the underlying LLM. Each agent role was initialized with the same role description and initial instructions across the two system configurations to ensure a fair comparison.

An agent selection function iterates through each worker agent and the conversation delegation agent (CDA), ensuring that all agents are represented in the discussion. Once all agents have accepted a proposed solution, marked through the "ACCEPT" flag, the CDA emits a "FINALIZE" flag, prompting the Documentation Engineer Agent to produce the final data pipeline output. The full algorithm for finalisation and ordering
of agents is displayed in the Appendix \ref{apdx:algorithms} Algorithm \ref{Alg:Finalisation}.


    

\subsection{Evaluation Metrics}
To evaluate the quality of the data pipeline designs generated by our memory agents,
and to compare to the pipeline designs generated from default Autogen, we use an
LLM as a judge \citep{zheng2023-llmjudge} to score each pipeline design and provide a qualitative analysis to support these scores. We evaluated the multi-agent system performance under the following metrics:
\begin{itemize}
    \item Scalability: ability for the data pipeline to handle increasing data volumes or user loads. 
    \item Reliability: ability for the data pipeline to handle failures and ensure data integrity. 
    \item Usability: is there enough detail in the data pipeline design for developers to implement the design?
    \item Cost-effectiveness: balance between costs and benefits of each chosen component. 
    \item Documentation: how well-justified and documented is the choice of elements for the data pipeline?
\end{itemize}

The scalability and reliability metrics are chosen as core requirements in modern data
pipelines. Scalability reflects the ability to grow the pipeline to handle larger
volumes of data and users, while reliability ensures the pipeline is consistent
and fault-tolerant, both of which are crucial if the pipeline were to be deployed.
Usability and documentation metrics reflect the details and design decisions taken.
A strong design is not useful if it does not contain enough detail or is too
abstract to be practically implemented. Usability measures whether the output
designs are detailed and clear enough for engineering teams to implement.
Design decisions must be well-documented, with clear justifications and explanations
for each component, which reveals the reasoning behind the agents' choices. 
Finally, the cost-effectiveness metric evaluates whether the pipeline design has
considered and balanced the need for computation resources with the cost of
those resources.
Run-time metrics such as latency and throughput are not included in our evaluation metrics
as we only present the design of the data pipelines to be evaluated, and do not implement
the designs into code

\subsection{Data pipeline design performance}\label{sec:data-pipeline-design-performance}
\begin{figure}
    \centering
    \includegraphics[width=0.8\linewidth]{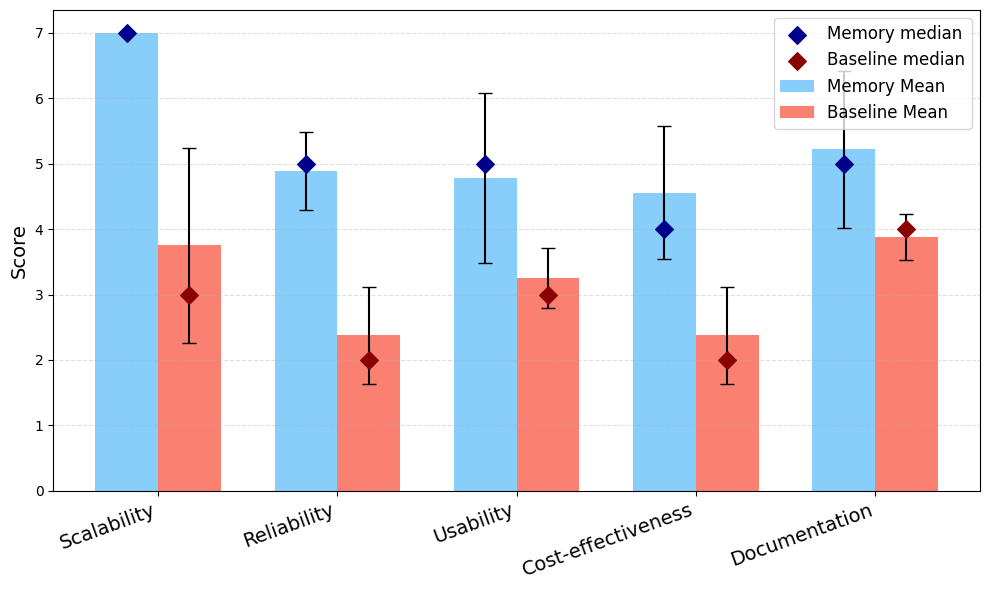}
    \caption{LLM-as-a-Judge metrics for the Data Pipeline design case study.}
    \label{fig:memory-stats-bars}
\end{figure}

The median and standard deviation of each quality metric is presented in
Figure \ref{fig:memory-stats-bars}. The Intrinsic Memory system
shows consistent improvement on all metrics compared to the baseline Autogen. 

The Documentation quality focuses on the clarity and how well-justified the
design choices are. While Intrinsic Memory helps to boost the Documentation score
over the baseline, the score is still relatively low at a mean of 4.9. This
suggests that retaining memory of the conversation alone does not guarantee
good justification, and while some context and attributes of each component
are remembered, the reasons for choosing the components are not. This could be a problem with the training corpus, and a requirement for better annotated training data. Similarly, the Usability score is low with means of 3.32 and 4.9 for the baseline and Intrinsic Memory, respectively. 

The improved quality comes at a cost of additional tokens outlined in Table \ref{tab:data-pipeline-metrics}. Intrinsic Memory
uses on average 32\% more tokens than the baseline as it outputs are more descriptive on average, although the number of conversation
turns is similar and not statistically significant. This indicates that the addition
of a memory module costs additional token overhead to maintain, but does not increase the number of conversation turns between agents.

\begin{table}[!tb]
\centering
\caption{Mean efficiency and LLM-as-a-Judge metrics after 10 independent runs,
with p-values calculated using a Wilcoxon ranked sum test.
Usability and number of conversation turns is
highlighted in italics as the metrics that do not show
statistical significance between the baseline and our Intrinsic Memory approach.}
\label{tab:data-pipeline-metrics}
\begin{tabular}{lp{1.5cm}p{1.5cm}p{1.2cm}}
\hline
\textbf{Metric} & \textbf{Baseline Autogen} & \textbf{Intrinsic Memory} & \textbf{p-value}\\
\hline
Tokens & 36077 & 47830 & 0.0195\\
\textit{Conversation turns} & 14.3 & 16 & \textit{0.2632}\\
\hline
Scalability & 3.75 & 7 & 0.0004 \\
Reliability & 2.37 & 4.9 & 0.0003\\
\textit{Usability} & 3.25 & 4.9 & 0.0093 \\
Cost-effectiveness & 2.37 & 4.7 & 0.001 \\
Documentation & 3.87 & 5.4 & 0.0077 \\
\hline
\end{tabular}
\end{table}

\subsection{Qualitative analysis of data pipeline outputs}

Figure \ref{fig:snippet-output} shows snippets for one component from the highest-scoring outputs for the intrinsic memory agent system and baseline Autogen system.

\begin{figure}[!tbh]
\centering

\begin{subfigure}[t]{0.95\linewidth}
\begin{tcolorbox}[colback=gray!10, colframe=gray!80,fontupper=\small\ttfamily]

"Component 1": "Data Ingestion (Amazon S3)"

\hspace{\parindent}"AWSName": "AmazonS3",

\hspace{\parindent}"Pros": ["Scalable", "durable", "secure storage for raw data"],

\hspace{\parindent}"Cons": ["Additional cost for storing large amounts of data"],

\hspace{\parindent}"Design": "Use S3 as a central repository for all data sources, with separate buckets for each source if needed.",

\hspace{\parindent}"Details": "Implement S3 event notifications to trigger processing workflows upon new data arrival."

\end{tcolorbox}
\caption{Intrinsic Memory Agent system sample from highest-scoring output. This data pipeline received scores of Scalability: 7, Reliability: 5, Usability: 6, Cost-effectiveness: 6, Documentation: 7}
\label{fig:intrinsic_memory_output_pipeline}
\end{subfigure}

\vspace{0.75em}

\begin{subfigure}[t]{0.95\linewidth}
\centering
\begin{tcolorbox}[colback=gray!10, colframe=gray!80,fontupper=\small\ttfamily]
"Component 1": {

\hspace{\parindent} "Name": "Data Ingestion",

\hspace{\parindent} "Description": "Ingest data from various sources (camera, lidar, radar) at high speeds",

\hspace{\parindent} "Implementation difficulties": 7, 

\hspace{\parindent} "Maintainability difficulties": 6
}
\end{tcolorbox}
\caption{Baseline Autogen system system sample from highest-scoring output. This data pipeline received scores of Scalability: 5, Reliability: 4, Usability: 3, Cost-effectiveness: 3, Documentation: 2}
\label{fig:baseline_autogen_output_pipeline}
\end{subfigure}

\caption{Snippets of one component within the data pipeline design from both systems. 
The full outputs can be found in the appendix in 
Figures \ref{fig:intrinsic_memory_output_pipeline_full}
and \ref{fig:baseline_autogen_output_pipeline_full}.}
\label{fig:snippet-output}
\end{figure}

The Intrinsic Memory Agent system outperforms the baseline system across the five quality metrics. 
In terms of scalability, the Intrinsic Memory Agent system is capable of providing an overall assessment of scalability, specifically around varying data volumes, whereas the baseline system encapsulates that measure only in the form of "maintenance difficulty" for each component of the pipeline.
In terms of reliability, the Intrinsic Memory Agent provides considerations for each component, such as AWS Kinesis's secure streaming capabilities and considerations as well as the use of Docker containers within amazon SageMaker to improve stability and reproducibility of ML pipelines.
The Intrinsic Memory Agent provides a more descriptive Usability output of the Intrinsic and a clearer pathway to implementation.
In terms of cost, the Intrinsic Memory Agent makes specific calculations and observations for the cost-effectiveness and resource requirements, including reasoning behind each component choice, whereas the baseline system limits itself to overall evaluations of implementation and maintainability difficulties.

Finally, the Intrinsic Memory Agent ultimately provides justification and documents its recommendation under each component, including pros and cons for each component choice.

Overall, the Intrinsic Memory Agent provides a more descriptive answer and more value to engineers by specifying tools, configurations and trade-offs. For example, its Data Streaming design recommends Amazon Kinesis, whereas the baseline simply states “Ingest data from various sources (camera, lidar, radar) at high speeds.” Similarly, the IMA cites the specific connections between components that must be implemented (for example, Amazon S3 => Amazon EC2 through API) . Although some precise configuration settings remain unspecified, the baseline merely names each component without offering implementation details or alternatives.

The components specified by the Intrinsic Memory Agent are more relevant to the problem specification. The data pipeline design task explicitly specifies the input data contains lidar and radar data sources, in which SageMaker and Kinesis are particularly relevant, specifically used for lidar data processing. This contrasts the vague "Lidar data processing" component of the baseline, which only contains a general description to process the data without providing any details.

\section{Discussion and Limitations}\label{sec:limitations}
Although the Intrinsic Memory Agent approach shows improved performance across the data pipeline generation task and the selected benchmarks, further validation is required across a broader set of complex tasks, potentially with varying number of agents, and models. Furthermore, our approach's performance and consistency comes at the cost of increased token usage, due to the additional update calls. Further work to reduce the number of update calls, updating only when necessary, will help to alleviate the additional usage.

The results demonstrate that a movement towards heterogeneity of agents leads to an improvement in performance of the multi-agent system, allowing agents to focus more specifically on an area of the design. This indicates that methods to provide additional heterogeneity, such as the ability to fine-tune agents towards their specialisation, might see additional performance gains, alongside the personalization of memories focused on individual experience. 

\section{Conclusion}\label{sec:conclusion}
This paper introduces Intrinsic Memory Agents, a novel multi-agent LLM framework that constructs agent-specific heterogeneous memories to enhance multi-agent collaboration in discussion and planning tasks. Evaluation PDDL dataset and on a practical data pipeline design problem demonstrates our framework's improved performance on structured planning tasks, with a 15.5\% increase over the next best memory architecture, at the cost of increased token usage.
Further evaluation on the ALFWorld and FEVER datasets demonstrates our approach's
consistent performance, ranking among the top memory mechanisms with the lowest standard deviation
while the best memory mechanisms on these benchmarks show high standard deviation.
Our strong performance using a generic memory template demonstrates the generalisablity of our
approach to other problems, without the need to hand-craft high quality memory templates. 

Results on the data pipeline case study further show the Intrinsic Memory Agents' enhanced ability to collaborate on complex tasks. The Intrinsic Memory Agent system outperforms the baseline system across all quality measures of scalability, reliability, usability, cost-effectiveness, and documentation, as well as an ability to more closely follow the task specification, providing more actionable recommendations by suggesting specific tools and frameworks, as well as trade-off details of each component in the pipeline. 

\newpage
\section*{Reproducibility Statement}

We have taken care to ensure that our experiments and results are transparent and reproducible by detailing the models, computational setup, code, statistical tests, and prompts used in our experiments. The LLM model (Llama3.2:3b) is cited, named, and referenced in the main text. The computational infrastructure used, including the GPU model names and operating system are specified in section \ref{sec:pddl-benchmark} of the main text. For code, the names and versions of relevant Python libraries are specified within the supplementary code files. A Wilcoxon rank-sum test is used to test statistical significance for the data pipeline case study. P-values and standard deviation measures are included in the performance analysis in section \ref{sec:data-pipeline-design-performance}. 5 independent runs with different set seed are used for the numeric benchmarks, with the seeds specified within the supplementary code. All code for running the data pipeline case study and numeric benchmarks are included in the supplementary materials. Finally, the selected prompts of the multi-agent and Intrinsic memory architecture are shown in Appendix \ref{apdx:data-pipeline-prompts}. Further prompts for each agent can be found as part of the supplementary code, under the ``prompts" directory.

\bibliography{iclr2026_conference}

@article{huang2025-hallucination-survey,
author = {Huang, Lei and Yu, Weijiang and Ma, Weitao and Zhong, Weihong and Feng, Zhangyin and Wang, Haotian and Chen, Qianglong and Peng, Weihua and Feng, Xiaocheng and Qin, Bing and Liu, Ting},
title = {A Survey on Hallucination in Large Language Models: Principles, Taxonomy, Challenges, and Open Questions},
year = {2025},
issue_date = {March 2025},
publisher = {Association for Computing Machinery},
address = {New York, NY, USA},
volume = {43},
number = {2},
issn = {1046-8188},
url = {https://doi.org/10.1145/3703155},
doi = {10.1145/3703155},
journal = {ACM Trans. Inf. Syst.},
month = jan,
articleno = {42},
numpages = {55}
}

@article{lewis2020retrieval,
  title={Retrieval-augmented generation for knowledge-intensive nlp tasks},
  author={Lewis, Patrick and Perez, Ethan and Piktus, Aleksandra and Petroni, Fabio and Karpukhin, Vladimir and Goyal, Naman and K{\"u}ttler, Heinrich and Lewis, Mike and Yih, Wen-tau and Rockt{\"a}schel, Tim and others},
  journal={Advances in Neural Information Processing Systems},
  volume={33},
  pages={9459--9474},
  year={2020}
}

@inproceedings{asai2024selfrag,
title={Self-{RAG}: Learning to Retrieve, Generate, and Critique through Self-Reflection},
author={Akari Asai and Zeqiu Wu and Yizhong Wang and Avirup Sil and Hannaneh Hajishirzi},
booktitle={The 12th International Conference on Learning Representations},
year={2024},
url={https://openreview.net/forum?id=hSyW5go0v8}
}

@misc{zhang2024surveymemoryllm,
      title={A Survey on the Memory Mechanism of Large Language Model based Agents}, 
      author={Zeyu Zhang and Xiaohe Bo and Chen Ma and Rui Li and Xu Chen and Quanyu Dai and Jieming Zhu and Zhenhua Dong and Ji-Rong Wen},
      year={2024},
      eprint={2404.13501},
      archivePrefix={arXiv},
      primaryClass={cs.AI},
      url={https://arxiv.org/abs/2404.13501}, 
}

@misc{guo2024llm-multiagents-survey,
      title={Large Language Model based Multi-Agents: A Survey of Progress and Challenges}, 
      author={Taicheng Guo and Xiuying Chen and Yaqi Wang and Ruidi Chang and Shichao Pei and Nitesh V. Chawla and Olaf Wiest and Xiangliang Zhang},
      year={2024},
      eprint={2402.01680},
      archivePrefix={arXiv},
      primaryClass={cs.CL},
      url={https://arxiv.org/abs/2402.01680}, 
}

@article{Li2024-multi-agent-survey,
  author    = {Xinyi Li and Sai Wang and Siqi Zeng and Yu Wu and Yi Yang},
  title     = {A survey on LLM-based multi-agent systems: workflow, infrastructure, and challenges},
  journal   = {Vicinagearth},
  volume    = {1},
  number    = {1},
  pages     = {9},
  year      = {2024},
  month     = {October},
  doi       = {10.1007/s44336-024-00009-2},
  url       = {https://doi.org/10.1007/s44336-024-00009-2},
  issn      = {3005-060X},
}

@misc{park2023generativeagents,
      title={Generative Agents: Interactive Simulacra of Human Behavior}, 
      author={Joon Sung Park and Joseph C. O'Brien and Carrie J. Cai and Meredith Ringel Morris and Percy Liang and Michael S. Bernstein},
      year={2023},
      eprint={2304.03442},
      archivePrefix={arXiv},
      primaryClass={cs.HC},
      url={https://arxiv.org/abs/2304.03442}, 
}

@misc{qian2025scalingmulti-agent,
      title={Scaling Large Language Model-based Multi-Agent Collaboration}, 
      author={Chen Qian and Zihao Xie and YiFei Wang and Wei Liu and Kunlun Zhu and Hanchen Xia and Yufan Dang and Zhuoyun Du and Weize Chen and Cheng Yang and Zhiyuan Liu and Maosong Sun},
      year={2025},
      eprint={2406.07155},
      archivePrefix={arXiv},
      primaryClass={cs.AI},
      url={https://arxiv.org/abs/2406.07155}, 
}

@misc{hong2024metagpt,
      title={MetaGPT: Meta Programming for A Multi-Agent Collaborative Framework}, 
      author={Sirui Hong and Mingchen Zhuge and Jiaqi Chen and Xiawu Zheng and Yuheng Cheng and Ceyao Zhang and Jinlin Wang and Zili Wang and Steven Ka Shing Yau and Zijuan Lin and Liyang Zhou and Chenyu Ran and Lingfeng Xiao and Chenglin Wu and Jürgen Schmidhuber},
      year={2024},
      eprint={2308.00352},
      archivePrefix={arXiv},
      primaryClass={cs.AI},
      url={https://arxiv.org/abs/2308.00352}, 
}

@misc{gao2024rag-survey,
      title={Retrieval-Augmented Generation for Large Language Models: A Survey}, 
      author={Yunfan Gao and Yun Xiong and Xinyu Gao and Kangxiang Jia and Jinliu Pan and Yuxi Bi and Yi Dai and Jiawei Sun and Meng Wang and Haofen Wang},
      year={2024},
      eprint={2312.10997},
      archivePrefix={arXiv},
      primaryClass={cs.CL},
      url={https://arxiv.org/abs/2312.10997}, 
}

@misc{huang2024-planning-agents,
      title={Understanding the planning of LLM agents: A survey}, 
      author={Xu Huang and Weiwen Liu and Xiaolong Chen and Xingmei Wang and Hao Wang and Defu Lian and Yasheng Wang and Ruiming Tang and Enhong Chen},
      year={2024},
      eprint={2402.02716},
      archivePrefix={arXiv},
      primaryClass={cs.AI},
      url={https://arxiv.org/abs/2402.02716}, 
}

@misc{yu2024-autorag,
      title={Auto-RAG: Autonomous Retrieval-Augmented Generation for Large Language Models}, 
      author={Tian Yu and Shaolei Zhang and Yang Feng},
      year={2024},
      eprint={2411.19443},
      archivePrefix={arXiv},
      primaryClass={cs.CL},
      url={https://arxiv.org/abs/2411.19443}, 
}

@misc{openai2024-gpt4ocard,
      title={GPT-4o System Card}, 
      author={OpenAI},
      year={2024},
      eprint={2410.21276},
      archivePrefix={arXiv},
      primaryClass={cs.CL},
      url={https://arxiv.org/abs/2410.21276}, 
}

@misc{anthropic-claude35,
  title        = {Claude 3.5 Sonnet},
  author       = {{Anthropic}},
  howpublished = {Technical announcement},
  year         = {2024},
  note         = {Available via Claude.ai and API},
  url          = {https://www.anthropic.com/news/claude-3-5-sonnet},
}

@misc{comanici2025-gemini25,
      title={Gemini 2.5: Pushing the Frontier with Advanced Reasoning, Multimodality, Long Context, and Next Generation Agentic Capabilities}, 
      author={Gheorghe Comanici and Eric Bieber and Mike Schaekermann and Ice Pasupat and Noveen Sachdeva and Inderjit Dhillon and Marcel Blistein and Ori Ram and Dan Zhang and Evan Rosen and Luke Marris and Sam Petulla and Colin Gaffney and Asaf Aharoni and Nathan Lintz and Tiago Cardal Pais and Henrik Jacobsson and Idan Szpektor and Nan-Jiang Jiang and Krishna Haridasan and Ahmed Omran and Nikunj Saunshi and Dara Bahri and Gaurav Mishra and Eric Chu and Toby Boyd and Brad Hekman and Aaron Parisi and Chaoyi Zhang and Kornraphop Kawintiranon and Tania Bedrax-Weiss and Oliver Wang and Ya Xu and Ollie Purkiss and Uri Mendlovic and Ilaï Deutel and Nam Nguyen and Adam Langley and Flip Korn and Lucia Rossazza and Alexandre Ramé and Sagar Waghmare and Helen Miller and Vaishakh Keshava and Ying Jian and Xiaofan Zhang and Raluca Ada Popa and Kedar Dhamdhere and Blaž Bratanič and Kyuyeun Kim and Terry Koo and Ferran Alet and Yi-ting Chen and Arsha Nagrani and Hannah Muckenhirn and Zhiyuan Zhang and Corbin Quick and Filip Pavetić and Duc Dung Nguyen and Joao Carreira and Michael Elabd and Haroon Qureshi and Fabian Mentzer and Yao-Yuan Yang and Danielle Eisenbud and Anmol Gulati and Ellie Talius and Eric Ni and Sahra Ghalebikesabi and Edouard Yvinec and Alaa Saade and Thatcher Ulrich and Lorenzo Blanco and Dan A. Calian and Muhuan Huang and Aäron van den Oord and Naman Goyal and Terry Chen and Praynaa Rawlani and Christian Schallhart and Swachhand Lokhande and Xianghong Luo and Jyn Shan and Ceslee Montgomery and Victoria Krakovna and Federico Piccinini and Omer Barak and Jingyu Cui and Yiling Jia and Mikhail Dektiarev and Alexey Kolganov and Shiyu Huang},
      year={2025},
      eprint={2507.06261},
      archivePrefix={arXiv},
      primaryClass={cs.CL},
      url={https://arxiv.org/abs/2507.06261}, 
}

@misc{li2024-longcontextllmsstruggle,
      title={Long-context LLMs Struggle with Long In-context Learning}, 
      author={Tianle Li and Ge Zhang and Quy Duc Do and Xiang Yue and Wenhu Chen},
      year={2024},
      eprint={2404.02060},
      archivePrefix={arXiv},
      primaryClass={cs.CL},
      url={https://arxiv.org/abs/2404.02060}, 
}

@misc{wang2025-recursivelysummarizing,
      title={Recursively Summarizing Enables Long-Term Dialogue Memory in Large Language Models}, 
      author={Qingyue Wang and Yanhe Fu and Yanan Cao and Shuai Wang and Zhiliang Tian and Liang Ding},
      year={2025},
      eprint={2308.15022},
      archivePrefix={arXiv},
      primaryClass={cs.CL},
      url={https://arxiv.org/abs/2308.15022}, 
}

@inproceedings{xu-2022-beyond-goldfish-memory,
    title = "Beyond Goldfish Memory: Long-Term Open-Domain Conversation",
    author = "Xu, Jing  and
      Szlam, Arthur  and
      Weston, Jason",
    editor = "Muresan, Smaranda  and
      Nakov, Preslav  and
      Villavicencio, Aline",
    booktitle = "Proceedings of the 60th Annual Meeting of the Association for Computational Linguistics (Volume 1: Long Papers)",
    month = may,
    year = "2022",
    address = "Dublin, Ireland",
    publisher = "Association for Computational Linguistics",
    url = "https://aclanthology.org/2022.acl-long.356/",
    doi = "10.18653/v1/2022.acl-long.356",
    pages = "5180--5197",
}

@misc{xu2025-a-mem,
      title={A-MEM: Agentic Memory for LLM Agents}, 
      author={Wujiang Xu and Kai Mei and Hang Gao and Juntao Tan and Zujie Liang and Yongfeng Zhang},
      year={2025},
      eprint={2502.12110},
      archivePrefix={arXiv},
      primaryClass={cs.CL},
      url={https://arxiv.org/abs/2502.12110}, 
}

@article{mem0,
  title={Mem0: Building Production-Ready AI Agents with Scalable Long-Term Memory},
  author={Chhikara, Prateek and Khant, Dev and Aryan, Saket and Singh, Taranjeet and Yadav, Deshraj},
  journal={arXiv preprint arXiv:2504.19413},
  year={2025}
}

@inproceedings{thorne-etal-2018-fever,
    title = "{FEVER}: a Large-scale Dataset for Fact Extraction and {VER}ification",
    author = "Thorne, James  and
      Vlachos, Andreas  and
      Christodoulopoulos, Christos  and
      Mittal, Arpit",
    editor = "Walker, Marilyn  and
      Ji, Heng  and
      Stent, Amanda",
    booktitle = "Proceedings of the 2018 Conference of the North {A}merican Chapter of the Association for Computational Linguistics: Human Language Technologies, Volume 1 (Long Papers)",
    month = jun,
    year = "2018",
    address = "New Orleans, Louisiana",
    publisher = "Association for Computational Linguistics",
    url = "https://aclanthology.org/N18-1074/",
    doi = "10.18653/v1/N18-1074",
    pages = "809--819",
}

@misc{ma2024-agentboard,
      title={AgentBoard: An Analytical Evaluation Board of Multi-turn LLM Agents}, 
      author={Chang Ma and Junlei Zhang and Zhihao Zhu and Cheng Yang and Yujiu Yang and Yaohui Jin and Zhenzhong Lan and Lingpeng Kong and Junxian He},
      year={2024},
      eprint={2401.13178},
      archivePrefix={arXiv},
      primaryClass={cs.CL}
}

@inproceedings{ALFWorld20,
  title ={{ALFWorld: Aligning Text and Embodied
           Environments for Interactive Learning}},
  author={Mohit Shridhar and Xingdi Yuan and
          Marc-Alexandre C\^ot\'e and Yonatan Bisk and
          Adam Trischler and Matthew Hausknecht},
  booktitle = {Proceedings of the International Conference on Learning Representations (ICLR)},
  year = {2021},
  url = {https://arxiv.org/abs/2010.03768}
}

@misc{zhang2025-gmemory,
      title={G-Memory: Tracing Hierarchical Memory for Multi-Agent Systems}, 
      author={Guibin Zhang and Muxin Fu and Guancheng Wan and Miao Yu and Kun Wang and Shuicheng Yan},
      year={2025},
      eprint={2506.07398},
      archivePrefix={arXiv},
      primaryClass={cs.MA},
      url={https://arxiv.org/abs/2506.07398}, 
}

@misc{zheng2023-llmjudge,
      title={Judging LLM-as-a-Judge with MT-Bench and Chatbot Arena}, 
      author={Lianmin Zheng and Wei-Lin Chiang and Ying Sheng and Siyuan Zhuang and Zhanghao Wu and Yonghao Zhuang and Zi Lin and Zhuohan Li and Dacheng Li and Eric P. Xing and Hao Zhang and Joseph E. Gonzalez and Ion Stoica},
      year={2023},
      eprint={2306.05685},
      archivePrefix={arXiv},
      primaryClass={cs.CL},
      url={https://arxiv.org/abs/2306.05685}, 
}

@misc{packer2024-memgpt,
      title={MemGPT: Towards LLMs as Operating Systems}, 
      author={Charles Packer and Sarah Wooders and Kevin Lin and Vivian Fang and Shishir G. Patil and Ion Stoica and Joseph E. Gonzalez},
      year={2024},
      eprint={2310.08560},
      archivePrefix={arXiv},
      primaryClass={cs.AI},
      url={https://arxiv.org/abs/2310.08560}, 
}

@misc{chen2024-compressivememory,
      title={Compress to Impress: Unleashing the Potential of Compressive Memory in Real-World Long-Term Conversations}, 
      author={Nuo Chen and Hongguang Li and Juhua Huang and Baoyuan Wang and Jia Li},
      year={2024},
      eprint={2402.11975},
      archivePrefix={arXiv},
      primaryClass={cs.CL},
      url={https://arxiv.org/abs/2402.11975}, 
}

@misc{li2024agentsneed,
      title={More Agents Is All You Need}, 
      author={Junyou Li and Qin Zhang and Yangbin Yu and Qiang Fu and Deheng Ye},
      year={2024},
      eprint={2402.05120},
      archivePrefix={arXiv},
      primaryClass={cs.CL},
      url={https://arxiv.org/abs/2402.05120}, 
}

@misc{he2025llmbasedmultiagentsystemssoftware,
      title={LLM-Based Multi-Agent Systems for Software Engineering: Literature Review, Vision and the Road Ahead}, 
      author={Junda He and Christoph Treude and David Lo},
      year={2025},
      eprint={2404.04834},
      archivePrefix={arXiv},
      primaryClass={cs.SE},
      url={https://arxiv.org/abs/2404.04834}, 
}
\bibliographystyle{iclr2026_conference}

\newpage
\appendix

\section{Algorithms}\label{apdx:algorithms}

This section contains the context construction algorithm presented in Section \ref{sec:intrinsic-memory} and the finalisation algorithm presented in \ref{sec:system_configuration}.

\newsavebox{\mycode}
\begin{lrbox}{\mycode}
\begin{lstlisting}[language=Python, xleftmargin=0pt]
def construct_context(conversation_history,
                      agent_memory, 
                      max_tokens):
context = []

# Include the initial task description
context.append(conversation_history[0])
context.append(agent_memory)

# Add most recent conversation turns until context limit is reached
remaining_tokens = max_tokens - count_tokens(context)
recent_turns = []

for turn in reversed(conversation_history[1:]):
    turn_tokens = count_tokens(turn)
    if turn_tokens <= remaining_tokens:
        recent_turns.insert(0, turn)
        remaining_tokens -= turn_tokens
    else:
        break

context.extend(recent_turns)
return context
\end{lstlisting}
\end{lrbox}

\begin{algorithm}
\caption{The context construction algorithm, which takes the current conversation history,
memory of the agent, and maximum number of tokens. It appends the most recent conversation
turn and agent memory to the context first, before using the remainder of the tokens
to append the rest of the conversation history, ensuring the memory and most recent
output is always included.}
\label{Alg:Context_construction}
\usebox{\mycode}
\end{algorithm}

\newsavebox{\finalisationcode}
\begin{lrbox}{\finalisationcode}
\begin{lstlisting}[language=Python]
workers = [BOA, DEA, MLA, IA]
global turn_counter
turn_counter += 1

if "FINALIZATION" in groupchat.messages[-1]['content']:
    return DJE
if last_speaker is CDA: 
    global worker_counter
    w = workers[worker_counter%4]
    worker_counter += 1
    print(f'worker_counter: {worker_counter}')
    return w
elif last_speaker in workers:
    if worker_counter %4 == 0:
        return KIA
    else: 
        return CDA
elif last_speaker is KIA:
        return ERA
elif last_speaker is ERA:
        return CDA
    
\end{lstlisting}
\end{lrbox}

\begin{algorithm}
\caption{The finalisation algorithm that specifies the order of agents speaking. It is a modified round-robin discussion between the agents: The discussion begins with each of the worker agents (BOA, DEA, MLA, IA) contributing to the conversation, with each worker's turn being followed by the conversation delegation agent (CDA). Once the workers have each had their turn, the knowledge integration agent and evaluation agent make their contributions, and the cycle begins again. The CDA is programmed to dedicate a certain number of turns to discussion, proposals, and consensus. The number of turns dedicated to each conversation stage is tracked, and once the consensus round is reached, each agent is asked to confirm if they agree with the proposed solution or not. If all agree on the proposed solution as being acceptable, the CDA will emit a "FINALIZE" response, triggering the documentation joining agent (DJE) to compile the agreed response and format it according to the task requirements.}
\label{Alg:Finalisation}
\usebox{\finalisationcode}
\end{algorithm}

\newpage

\section{Ablation study}\label{apdx:ablation}

We conduct an ablation study to understand the sensitivity of the
Intrinsic Memory Agents to the structure of the templates that the agents
use to update their memory.
We evaluate three approaches:
\begin{enumerate}
    \item \textbf{Manual template} - Manual hand-crafted templates for each type of dataset and problem, based on the kind of task the agents have to solve.
    \item \textbf{Generic template} - A generic template to store updated memory without any reference to specific problems or fields.
    \item \textbf{LLM-generated template} - We prompt the agent LLM to self-generate a template based on the instructions of the task, allowing the agent to dynamically create templates it deems suitable for each task.
\end{enumerate}

The approaches are evaluated on the PDDL, FEVER, and ALFWorld benchmark problems, tested across different base language models with multiple independent runs using set seeds for reproducibility and consistency.

\subsection{Template approaches}

\subsubsection{Manual template}
In the manual template approach, a strict structured template is used that agents must follow
to update their internal memory. The manual templates are created to track the relevant
information of the given task, with specific fields.
\begin{figure*}[!tbh]
\centering
\begin{tcolorbox}[colback=gray!10, colframe=gray!80,fontupper=\footnotesize\ttfamily]
You are a MEMORY UPDATER for a PDDL-style planning agent.

Your job:
- Maintain a compact JSON memory capturing stable, reusable information across tasks and domains.
- Only store information that improves future planning: common strategies, mistakes, valid action patterns, state-transition insights.
- Do not store long histories. Keep everything concise and deduplicated.

Inputs you receive each update call:
- currentmemory: the previous memory as json (may be empty re-init).
- latestturn: the agent’s most recent Thought/Action/Observation.
- currenttask: one of {blockworld, barman, gripper, tyreworld}.
- goal: current goal description.

OUTPUT:
- Return ONLY the updated memory as valid JSON following the template below.
- No extra commentary.

--------------------------
MEMORY TEMPLATE (ALWAYS FOLLOW)

{
  "task summary": "brief description of PDDL planning setting",
  "global strategies": [
    "high-level reusable planning heuristics across domains"
  ],
  "domains": {
    "blockworld": {
      "valid action patterns": ["pickup X", "putdown X", "stack X Y", "unstack X Y"],
      "good strategies": ["free target block before stacking"],
      "invalid patterns": ["wrong think format", "stack without clear base"],
      "mistakes": ["attempting pickup while arm full"]
    },
    "barman": {
      "valid action patterns": ["hand grasp glass", "fill-shot ...", "pour-shot-to-clean-shaker ..."],
      "good strategies": ["ensure hand availability before filling"],
      "invalid patterns": ["fill without holding glass"],
      "mistakes": ["grasp with occupied hand"]
    },
    "gripper": {
      "valid action patterns": ["move R1 R2", "pick O Room Gripper", "drop O Room Gripper"],
      "good strategies": ["carry multiple items before moving rooms"],
      "invalid patterns": ["drop object in wrong room"],
      "mistakes": ["pick while gripper full"]
    },
    "tyreworld": {
      "valid action patterns": ["open X", "fetch O C", "loosen N H", "jack-up H"],
      "good strategies": ["open boot early to access tools"],
      "invalid patterns": ["loosen nut without wrench"],
      "mistakes": ["inflate wheel without pump"]
    }
  },
  "tasks": [
    {
      "id": "identifier or hash of goal",
      "goal": "exact goal text",
      "status": "pending|solved",
      "helpful observations": ["short state insights from valid steps"],
      "invalid actions": ["summaries of failed attempts"],
      "progress notes": ["short planning insights for this task"]
    }
  ]
}

--------------------------
UPDATE INSTRUCTIONS

1. Parse current memory.  
   - If empty or invalid, initialize using the template above.

2. Update the domain-specific sections:
   - From latest turn, add new useful action patterns, invalid patterns, or mistakes.
   - Keep lists short, deduplicated, and generalisable.

3. Update global strategies if the latest turn reveals a robust cross-domain heuristic.

4. Update the relevant task entry:
   - If no entry exists for this goal, create one.
   - Add helpful observations if new actionable state insights appear.
   - Add invalid actions if latest turn shows an invalid move.
   - Add progress notes for general reasoning improvements.
   - If task finished, mark status = "solved".

5. Return ONLY the updated JSON memory, nothing else.

\end{tcolorbox}
\caption{Manually generated prompt for the LLM agent in the PDDL task.}
\label{fig:intrinsicmemory-manual-template-prompt}
\end{figure*}

\subsubsection{Generic template}
For the generic approach, we use a universal template structure, giving only general fields to be filled in with
the current task description and trajectory, allowing the model to use any form of memory representation, and
not strictly requiring the same structured format to be used each time. 

\begin{figure*}[!tbh]
\centering
\begin{tcolorbox}[colback=gray!10, colframe=gray!80,fontupper=\footnotesize\ttfamily]
Use your latest response to populate and update the current memory with factual information to solve the task 
based on the task description.\\

\#\# Task Description:\\

\{task\_description\}\\

\#\# Current Task Trajectory:\\

\{task\_trajectory\}\\

\#\# Current Memory:\\

\{current\_memory\}\\

\end{tcolorbox}
\caption{Prompt for the LLM agent to update its memory}
\label{fig:intrinsicmemory-llm-notemplate-prompt}
\end{figure*}

\subsubsection{LLM-generated template}
Finally, we leverage each agent's own capabilities to create a template during initialisation which it deems suitable for the task at hand. This approach removes the need to create manual templates for each type of task while allowing for memory templates specific to the task. The template generation prompt is as follows:
\begin{figure*}[!tbh]
\centering
\begin{tcolorbox}[colback=gray!10, colframe=gray!80,fontupper=\footnotesize\ttfamily]
I have an AI agent that has to complete a task. 

The agent has a memory that is updated each time the LLM responds by comparing the latest response and the existing memory, and adding any new important information. The memory should be templated based on the nature of the task following a json-style format. 
The memory update is conducted as a prompted LLM call to update the memory. Provide the instructions to the agent for such an update operation, as well as the generic memory template for this particular task. Provide the full answer as a single prompt. 
Only include the most crucial details to the updating instructions to preserve token usage. 
Do not explain or describe the prompt, simply return the prompt and nothing more. 

This is the task description:
\{task\_description\}

\end{tcolorbox}
\caption{Prompt for the LLM agent to generate its own memory template.}
\label{fig:intrinsicmemory-llm-template-prompt}
\end{figure*}

\subsection{Ablation results}
We perform the ablation study on two additional models: Gemma3-12b and Mistral-7b. 
Tables \ref{tbl:mistral-ablation} and \ref{tbl:gemma-ablation} display the mean and
standard deviation of each template approach on the PDDL, FEVER, and ALFWorld
benchmarks. We find that the approaches using an LLM-generated template or
a generic template outperform the manual template approach. In general, the generic
template performs best, but performance for both the LLM-generated approach
and generic approach are similar using Gemma3. Token usage varies based on the underlying
model. For example, Mistral uses more tokens on average for the generic approach whereas
Gemma uses more tokens for the LLM-based approach. Note that the cost of the template generation
in the LLM approach is a fixed one-time cost at the beginning of each problem, and therefore
adds little overhead compared to the rest of the runtime.
\begin{table}[!tbh]
    \centering
    \begin{tabular}{lllll}
    \hline
    \multirow{2}{*}{\textbf{Benchmark}} & \multirow{2}{*}{\textbf{Memory}} & \multicolumn{2}{c}{\textbf{Rewards}} & \multirow{2}{*}{\textbf{Average tokens}} \\
    & & Mean & Std & \\ \hline

    \multirow{3}{*}{PDDL} & Intrinsic & 0.063648 & 0.004113 & 848,052 \\
                          & Intrinsic-LLM & \textbf{0.069740} & 0.010224 & 609,102 \\
                          & Intrinsic-Generic & 0.066198 & 0.008096 & 613,631 \\ \hline
                          
    \multirow{3}{*}{FEVER} & Intrinsic & 0.119974 & 0.024228 & 1,395,395 \\
                           & Intrinsic-LLM & 0.300746 & 0.051672 & 949,377 \\
                           & Intrinsic-Generic & \textbf{0.379853} & 0.041365 & 1,079,041 \\ \hline
                          
    \multirow{3}{*}{ALFWorld} & Intrinsic & 0.015390 & 0.000317 & 1,789,989 \\
                              & Intrinsic-LLM & 0.014944 & 0.000046 & 991,884 \\
                              & Intrinsic-Generic & \textbf{0.029926} & 0.000116 & 1,027,602 \\ \hline
    \end{tabular}
    \caption{Mistral:7b}
    \label{tbl:mistral-ablation}
\end{table}

\begin{table}[!tbh]
    \centering
    \begin{tabular}{lllll}
    \hline
    \multirow{2}{*}{\textbf{Benchmark}} & \multirow{2}{*}{\textbf{Memory}} & \multicolumn{2}{c}{\textbf{Rewards}} & \multirow{2}{*}{\textbf{Average tokens}} \\
    & & Mean & Std & \\ \hline

    \multirow{3}{*}{PDDL} & Intrinsic & 0.253164 & 0.049818 & 379,786 \\
                          & Intrinsic-LLM & 0.253986 & 0.032846 & 359,544 \\
                          & Intrinsic-Generic & \textbf{0.260255} & 0.022382 & 350,541 \\ \hline
                          
    \multirow{3}{*}{FEVER} & Intrinsic & 0.608603 & 0.006874 & 178,456 \\
                           & Intrinsic-LLM & 0.649391 & 0.004576 & 86,849 \\
                           & Intrinsic-Generic & \textbf{0.653500} & 0.005910 & 88,274 \\ \hline
                          
    \multirow{3}{*}{ALFWorld} & Intrinsic & 0.025065 & 0.000262 & 942,916 \\
                              & Intrinsic-LLM & 0.045002 & 0.000350 & 883,913 \\
                              & Intrinsic-Generic & \textbf{0.048296} & 0.008294 & 879,134 \\ \hline
    \end{tabular}
    \caption{Gemma3:12b}
    \label{tbl:gemma-ablation}
\end{table}

\clearpage
\newpage
\section{Full benchmarking results}\label{apdx:benchmarking-results}

\begin{table}[!tbh]
    \centering
    \begin{tabular}{lllll}
    \hline
    \multirow{2}{*}{\textbf{Benchmark}} & \multirow{2}{*}{\textbf{Memory}} & \multicolumn{2}{c}{\textbf{Rewards}} & \multirow{2}{*}{\textbf{Average tokens}} \\
    & & Mean & Std & \\ \hline
    \multirow{9}{*}{ALFWorld} 
     & ChatDev & 0.020904 & 0.006491 & 174,451 \\
     & Empty & 0.020408 & 0.000000 & 113,817 \\
     & G-Memory & 0.040851 & 0.019529 & 104,492 \\
     & Generative & 0.060656 & 0.031094 & 181,211 \\
     & MemoryBank & 0.031623 & 0.004932 & 160,273 \\
     & MetaGPT & 0.030777 & 0.000547 & 136,716 \\
     & Voyager & \textbf{0.072112} & \textbf{0.034554} & 102,482 \\
     & \textit{Intrinsic-Generic} & 0.048296 & 0.008294 & 784,119 \\
     & \textit{Intrinsic-LLM} & 0.045002 & 0.000350 & 882,158 \\
    \hline
    \multirow{9}{*}{FEVER} 
     & ChatDev & 0.617500 & 0.016583 & 80,656 \\
     & Empty & 0.646667 & 0.015706 & 55,539 \\
     & G-Memory & 0.628792 & 0.027763 & 267,823 \\
     & Generative & 0.651250 & 0.011087 & 117,278 \\
     & MemoryBank & 0.632839 & 0.010771 & 65,997 \\
     & MetaGPT & \textbf{0.667000} & \textbf{0.028853} & 48,998 \\
     & Voyager & 0.643000 & 0.025884 & 62,555 \\
     & \textit{Intrinsic-Generic} & 0.653500 & 0.005910 & 88,274 \\
     & \textit{Intrinsic-LLM} & 0.649391 & 0.004576 & 86,849 \\
    \hline
    \multirow{9}{*}{PDDL} 
     & ChatDev & 0.222746 & 0.021114 & 70,765 \\
     & Empty & 0.224329 & 0.019766 & 52,075 \\
     & G-Memory & 0.152222 & 0.039180 & 162,712 \\
     & Generative & 0.164944 & 0.021977 & 84,113 \\
     & MemoryBank & 0.158083 & 0.013697 & 66,299 \\
     & MetaGPT & 0.197106 & 0.018546 & 92,436 \\
     & Voyager & 0.191088 & 0.031562 & 68,483 \\
     & \textit{Intrinsic-Generic} & \textbf{0.260255} & \textbf{0.022382} & 352,301 \\
     & \textit{Intrinsic-LLM} & 0.253986 & 0.032846 & 359,302 \\
    \hline
    \end{tabular}
    \caption{Rewards and tokens for each memory framework on the three benchmark problem sets. }
    \label{tbl:gemma3-results-table}
\end{table}

\clearpage
\newpage

\section{Full prompts and example outputs}\label{apdx:data-pipeline-prompts}

\begin{figure}[!tbh]
    \centering
    \begin{tcolorbox}[colback=gray!10, colframe=gray!80,fontupper=\ttfamily]
    You are maintaining the memory of an agent working as [ROLE] in a multi-agent conversation.
    Use the old memory and the newest output by the agent to populate and up-
    date the current memory json with factual information. \\
    
    For context, old memory content: 
    
    [MEMORY\_CONTENT]\\
    
    Current content generated by the agent:
    
    [AGENT\_OUTPUT]\\
    
    Update the memory content to incorporate new information while preserving 
    key historical context. 
    The updated content should be concise and focus on information relevant to 
    both the old memory and the newly generated output.
    
    \end{tcolorbox}
    \caption{Prompt of the memory update function, where \textit{ROLE} is the agent's role specification $R_n$;  \textit{MEMORY\_CONTENT} is the current content $M_{n,m-1}$; \textit{AGENT\_OUTPUT} is the agent's output $O_{n,m}$.}
    \label{fig:memory-update-prompt}
\end{figure}

\begin{figure*}[!tbh]
\centering
\begin{tcolorbox}[colback=gray!10, colframe=gray!80,fontupper=\small\ttfamily]

```
This discussion session is set up to discuss the best data pipeline for a real time data intensive machine learning training and inference self driving application. The goal is to discuss and find consensus on how to set up the data pipeline, including each component in the data pipeline. 

You can assume that we have access to AWS. \\

**Data Description:**
Real-time data of cars driving in street. 

There are 6 camera sources with data in .jpg format; 1 lidar source in .pcd.bin format; and 5 radar sources with data in .pcd format. \\

**Discussion and Design:**

- Emphasize comprehensive understanding of the data sources, processing requirements, and desired outcomes.

- Encourage each other to engage in an open discussion on potential technologies, components, and architectures that can handle the diverse data streams and real-time nature of the data.

- Keep the conversation on design and evaluating the pros and cons of different design choices, considering scalability, maintainability, and cost-effectiveness.

- The team should agrees on a final architectural design, justifying the choices made.

- The team should produce the required the document PIPELINE\_OVERVIEW.json.\\

**Final Output:**

- Produce a concise summary of the agreed-upon pipeline architecture, highlighting its key components and connections.

- Provide a high-level plan and rationale for the design, explaining why it is well-suited for the given data and use case.

- Estimate the cloud resources, implementation efforts, and associated costs, providing a rough breakdown and complexity rating.

- Generate a `PIPELINE\_OVERVIEW.json` file, detailing the proposed complete architecture in JSON format with the following fields: 

 - “Platform“: A cloud service provider’s name if the cloud solution is the best, or “local server” if locally hosted servers are preferred. 
 
 - “Component 1”: The first component in the pipeline framework. 
 
 - “Component 2”: The second component in the pipeline framework. Continue until all required components are listed. 
 
 - “Implementation difficulties": A rating from 1 to 10 (lowest to highest).
 
 - “Maintenance difficulties”: A rating from 1 to 10 (lowest to highest). \\

**Instructions:**

- Remember, this is a collaborative design discussion, not a project execution. Refrain from assigning tasks with deadlines.

- Keep the conversation focused on architectural choices, technologies, and potential challenges.

- Emphasize the importance of a well-thought-out design.
```
\end{tcolorbox}
\caption{Prompt for the Data Pipeline design. This is the task
description provided to all agents in the system.}
\end{figure*}

\begin{figure*}[!th]
\centering
\begin{tcolorbox}[colback=gray!10, colframe=gray!80,fontupper=\ttfamily]

```

You are an expert in data pipeline design evaluation. Your task is to evaluate the following data pipeline design based on its description. For each of the specified metrics, assign a score from 1 to 10, where 1 is the lowest and 10 is the highest. Provide a brief justification for each score. Be critical and harsh if the design is poor and give it a low score. Base your evaluation solely on the provided description. Do not assume any additional information.\\
 
Metrics: 

1. Scalability: Ability to handle increasing data volumes or user loads.

2. Reliability: Ability to handle failures and ensure data integrity. 

3. Usability: Enough detail for developers to implement the design.

4. Cost-effectiveness: Balance between costs and benefits. 

5. Documentation: How well-justified is the choice of elements for the data pipeline\\
 
Provide your evaluation in the following format in a json dict: {

[metric1]: {{score}: [score], 

{justification}: [justification]}, 

[metric2]...

}

```
\end{tcolorbox}
\caption{Prompt for the LLM-as-a-Judge to evaluate data pipeline designs.}
\end{figure*}

\begin{figure*}[!th]
\centering
\begin{tcolorbox}[colback=gray!10, colframe=gray!80,fontupper=\small\ttfamily]

"Platform": "AWS"\\

"Component 1": "Amazon S3 (Data Storage)"\\
"pros": "Scalable, durable, and secure storage for raw data."\\
"cons": "Additional cost for storing large amounts of data."\\
"design": "Use S3 as a central repository for all data sources, with separate buckets for each source if needed."\\
"details": "Implement S3 event notifications to trigger data processing workflows upon new data arrival."\\

"Component 2": "Amazon EC2 (GPU Instances) — Image \& Point Cloud Processing"\\
"pros": "Scalable, high-performance computing resources for complex image and point cloud processing."\\
"cons": "Higher cost due to instance usage and GPU pricing."\\
"design": "Use EC2 NVIDIA GPU instances for camera source data processing."\\
"details": "Implement Docker-based containerization for reproducibility and efficient execution."\\

"Component 3": "Amazon SageMaker — Model Training \& Processing"\\
"pros": "Fully managed, high-performance infrastructure for ML training and inference."\\
"cons": "Higher cost for GPU-backed instances."\\
"design": "Use SageMaker for lidar and radar data processing, including preprocessing and feature extraction."\\
"details": "Use Docker containers to ensure stable, reproducible ML pipelines."\\

"Component 4": "Amazon Kinesis — Real-Time Streaming"\\
"pros": "Scalable, fast, and secure streaming and analytical capabilities."\\
"cons": "Additional cost for high-throughput streaming."\\
"design": "Use Kinesis for real-time ingestion, analytics, and event-driven triggers."\\
"details": "Implement aggregation, filtering, and visualization workflows to support decisions."\\

"Component 5": "Amazon Redshift — Batch Analytics"\\
"pros": "High-performance analytics engine with strong scaling capabilities."\\
"cons": "Additional cost for warehousing large datasets."\\
"design": "Use Redshift for batch analytical workloads across multi-source data."\\
"details": "Implement analytics queries and visualization pipelines for reporting and BI."\\
  
  
  
  
  
  
  
  

\end{tcolorbox}
\caption{Example output from intrinsic memory agent system which scored the highest. The intrinsic memory agent system adheres to the initial prompt well, providing an answer in the correct format, as well as providing sufficient justification for the choice of components, as well as implementation details.}
\label{fig:intrinsic_memory_output_pipeline_full}
\end{figure*}

\begin{figure*}[!th]
\centering
\begin{tcolorbox}[colback=gray!10, colframe=gray!80,fontupper=\small\ttfamily]
"Component 6": "Amazon IAM — Identity \& Access Control"\\
"pros": "Secure, scalable identity and permissions management."\\
"cons": "Requires careful design to avoid security misconfiguration."\\
"design": "Implement IAM roles and policies to manage access across all pipeline components."\\
"details": "Use Cognito for user authentication and authorization where appropriate."\\

"Component 7": "Amazon Cognito — Authentication"\\
"pros": "Managed, secure identity federation and authentication."\\
"cons": "Additional cost depending on usage tiers."\\
"design": "Use Cognito for managing user identity, login, and token issuance."\\
"details": ""\\
"Connections 1": "Amazon S3 → Amazon EC2 (GPU instances) (API)"\\
"Connections 2": "Amazon EC2 (GPU instances) → Amazon SageMaker (API)"\\
"Connections 3": "Amazon SageMaker → Amazon Kinesis (API)"\\
"Connections 4": "Amazon Kinesis → Amazon Redshift (API)"\\
"Connections 5": "Amazon IAM → Amazon Cognito (API)"\\

"Potential Challenge 1": "Scalability"\\
"Potential Challenge 2": "Performance"\\
"Potential Challenge 3": "Security"\\

"Rationale": "Pipeline is designed to be scalable, maintainable, and secure using microservices-based architecture."\\

"Complexity Rating": "7"\\

"Estimated Cloud Resource 1": "S3: 10–20 TB of storage"\\
"Estimated Cloud Resource 2": "EC2 (GPU instances): 4–8 GPU-enabled instances"\\
"Estimated Cloud Resource 3": "SageMaker: 2–4 GPU-enabled instances"\\
"Estimated Cloud Resource 4": "Kinesis: 2–4 streams"\\
"Estimated Cloud Resource 5": "Redshift: 2–4 clusters"\\

"Implementation Effort": "12–16 weeks"\\

"Cost 1": "S3: \$500–\$2,000 per month"\\
"Cost 2": "EC2 (GPU instances): \$5,000–\$20,000 per month"\\
"Cost 3": "SageMaker: \$2,000–\$10,000 per month"\\
"Cost 4": "Kinesis: \$1,000–\$5,000 per month"\\
"Cost 5": "Redshift: \$2,000–\$10,000 per month"\\

\end{tcolorbox}
\caption{Example output from intrinsic memory agent system which scored the highest. The intrinsic memory agent system adheres to the initial prompt well, providing an answer in the correct format, as well as providing sufficient justification for the choice of components, as well as implementation details. (Continued)}
\label{fig:intrinsic_memory_output_pipeline_full_2}
\end{figure*}

\begin{figure*}[!tbh]
\centering
\begin{tcolorbox}[colback=gray!10, colframe=gray!80,fontupper=\footnotesize\ttfamily]
{
  "Platform": "AWS",
  
  "Component 1": {
  
    "Name": "Data Ingestion",
    
    "Description": "Ingest data from various sources (camera, lidar, radar) at high speeds",
    
    "Implementation difficulties": 7,
    
    "Maintainability difficulties": 6
  },
  
  "Component 2": {
  
    "Name": "Image Processing",
    
    "Description": "Process camera data in real-time to extract relevant features",
    
    "Implementation difficulties": 8,
    
    "Maintainability difficulties": 7
    
  },
  
  "Component 3": {
  
    "Name": "Object Detection",
    
    "Description": "Detect objects of interest (e.g., pedestrians, cars, lanes) within image frames and track their movement over time",
    
    "Implementation difficulties": 9,
    
    "Maintainability difficulties": 8
    
  },
  
  "Component 4": {
  
    "Name": "Lidar Data Processing",
    
    "Description": "Process lidar data in real-time to estimate distances, velocities, and other relevant metrics",
    
    "Implementation difficulties": 8,
    
    "Maintainability difficulties": 7
    
  },
  
  "Component 5": {
  
    "Name": "Radar Data Processing",
    
    "Description": "Process radar data in real-time to estimate distances, velocities, and other relevant metrics",
    
    "Implementation difficulties": 8,
    
    "Maintainability difficulties": 7
    
  },
  
  "Component 6": {
  
    "Name": "Data Fusion",
    
    "Description": "Fuse the outputs from different sensors (camera, lidar, radar) to create a more accurate representation of the environment",
    
    "Implementation difficulties": 9,
    
    "Maintainability difficulties": 8
    
  },
  
  "Component 7": {
  
    "Name": "Model Training",
    
    "Description": "Train machine learning models on large datasets using AWS SageMaker's Training Grounds feature",
    
    "Implementation difficulties": 8,
    
    "Maintainability difficulties": 7
  },
  
  "Component 8": {
  
    "Name": "Inference",
    
    "Description": "Perform real-time inference on trained models, making predictions on new, unseen data",
    
    "Implementation difficulties": 9,
    
    "Maintainability difficulties": 8
    
  }
}

\end{tcolorbox}
\caption{Example output from baseline Autogen system which scored the highest.}
\label{fig:baseline_autogen_output_pipeline_full}
\end{figure*}

\end{document}